\documentclass[sigconf]{acmart}

\usepackage{booktabs} 
\usepackage{graphicx} 
\usepackage{subcaption} 
\usepackage{amsmath} 
\usepackage{listings} 
\usepackage{balance} 
\usepackage{hyperref} 
\usepackage{multirow} 
\usepackage{tfrupee} 

\copyrightyear{2025}
\acmYear{2025}
\setcopyright{none} 
\acmConference[xxxx]{Name}{July 2025}{Bengaluru, India}
\acmBooktitle{<Name>, July 2025, Bengaluru, India}
\acmDOI{--------------}
\acmISBN{----------------}

\renewcommand\footnotetext{} 

\title[A Hybrid ML Framework for Agricultural Recommendation]{A Hybrid Machine Learning Framework for Optimizing Crop Selection via Agronomic and Economic Forecasting}
\subtitle{A Voice-Enabled Decision Support System for Farmers in Karnataka}

\author{Niranjan Mallikarjun Sindhur}
\affiliation{%
  \institution{Department of Artificial Intelligence and Machine Learning \\ RV College of Engineering}
  \city{Bangalore}
  \state{Karnataka}
  \country{India}
}
\email{niranjanms.ai22@rvce.edu.in}

\author{Pavithra C}
\affiliation{%
  \institution{Department of Artificial Intelligence and Machine Learning \\ RV College of Engineering}
  \city{Bangalore}
  \state{Karnataka}
  \country{India}
}
\email{pavithrac.ai22@rvce.edu.in}

\author{Dr. Nivya Muchikel}
\affiliation{%
  \institution{Department of Mathematics \\ RV College of Engineering}
  \city{Bangalore}
  \state{Karnataka}
  \country{India}
}
\email{nivyamuchikel@rvce.edu.in}

\renewcommand{\shortauthors}{Niranjan M Sindhur et al.}

\begin{abstract}
\bfseries
Farmers in developing regions like Karnataka, India, face a dual challenge: navigating extreme market and climate volatility while being excluded from the digital revolution due to literacy barriers. This paper presents a novel decision support system that addresses both challenges through a unique synthesis of machine learning and human-computer interaction. We propose a hybrid recommendation engine that integrates two predictive models: a Random Forest classifier to assess agronomic suitability based on soil, climate, and real-time weather data, and a Long Short-Term Memory (LSTM) network to forecast market prices for agronomically viable crops. This integrated approach shifts the paradigm from "what can grow?" to "what is most profitable to grow?", providing a significant advantage in mitigating economic risk. The system is delivered through an end-to-end, voice-based interface in the local Kannada language, leveraging fine-tuned speech recognition and high-fidelity speech synthesis models to ensure accessibility for low-literacy users. Our results show that the Random Forest model achieves 98.5\% accuracy in suitability prediction, while the LSTM model forecasts harvest-time prices with a low margin of error. By providing data-driven, economically optimized recommendations through an inclusive interface, this work offers a scalable and impactful solution to enhance the financial resilience of marginalized farming communities.
\end{abstract}



\begin{document}
\maketitle

\section{Introduction}
The agricultural sector in Karnataka, India, a region where farming is the primary livelihood for a majority of the population, is at a precarious crossroads. Farmers are increasingly exposed to a confluence of systemic risks, including heightened climate variability, which leads to unpredictable yields, and extreme volatility in commodity markets, which decouples effort from financial reward \cite{dev2012small}. This economic precarity is exacerbated by a persistent information asymmetry, where farmers lack access to the timely, data-driven insights necessary for optimal decision-making.

In response, the field of digital agriculture has emerged, promising to empower farmers with technological solutions. However, the impact of these solutions has been severely limited by a significant digital literacy divide \cite{medero2021digital}. Despite the high penetration of smartphones in rural areas, many platforms are designed with a text-heavy, GUI-based approach that implicitly assumes a level of literacy that a large segment of the farming population does not possess. This creates a paradox where the tools for empowerment are available but not accessible.

This paper introduces a novel agricultural advisory system designed to systematically dismantle these intertwined barriers. Our central hypothesis is that an effective intervention must satisfy two conditions: it must provide recommendations that are both agronomically sound and economically optimal, and it must deliver this information through a medium that is inclusive and accessible to its target users.

To meet these conditions, we propose a system with two core technical contributions:
\begin{enumerate}
    \item \textbf{A Hybrid Agro-Economic Recommendation Engine:} We architect a two-stage predictive pipeline. The first stage employs a Random Forest classifier to determine crop suitability based on a rich feature set of soil nutrients (N, P, K, pH), long-term rainfall patterns, and real-time, localized weather data. The second stage takes the most suitable crops as input and utilizes a Long Short-Term Memory (LSTM) recurrent neural network to forecast their market prices at the expected time of harvest. This integration ensures that the final recommendation is optimized not just for yield, but also for profitability.

    \item \textbf{A Voice-First, Asymmetric Interaction Architecture:} To overcome the literacy barrier, the system is engineered around a voice-based interface that operates entirely in Kannada, the local language. We employ an "asymmetric" design, using a computationally efficient ASR model for transcription and a high-fidelity, hyper-localized TTS model for output synthesis.\\
    \hspace*{1em}Specifically, we use \texttt{distil-whisper/distil-large-v2} for ASR and \texttt{facebook/mms-tts-kan} for TTS. This prioritizes user comprehension and trust without sacrificing scalability.
\end{enumerate}

By synthesizing these approaches, we present an end-to-end decision support system that is technically robust, contextually aware, and socially inclusive. We provide a practical and scalable framework for leveraging advanced AI to generate tangible economic value for marginalized farming communities, thereby offering a pathway towards a more resilient and equitable agricultural future.

\section{Related Work}
Our research is positioned at the intersection of machine learning in agriculture and human-centered computing for development.

\subsection{Machine Learning for Crop Recommendation}
The application of machine learning to crop recommendation is an active area of research. Early approaches often relied on static rule-based systems, while more recent work has focused on supervised learning. Models such as Support Vector Machines (SVM), Naive Bayes, and particularly Random Forests have been widely adopted \cite{pudumalar2016crop}. Random Forest classifiers are favored for their robustness to overfitting, their ability to handle high-dimensional and non-linear data, and their inherent capacity to rank feature importance \cite{breiman2001random}. However, a significant limitation of most existing models is their exclusive focus on agronomic parameters. They excel at predicting which crops will grow well but fail to consider whether those crops will be profitable, a critical oversight that our work directly addresses.

\subsection{Time-Series Forecasting in Agricultural Markets}
Forecasting agricultural commodity prices is a challenging task due to the complex interplay of factors like seasonality, weather events, government policies, and market sentiment. Traditional econometric models like ARIMA (Autoregressive Integrated Moving Average) have been used but often struggle to capture the deep non-linearities present in the data \cite{paul2014arima}. With the rise of deep learning, Recurrent Neural Networks (RNNs), and specifically Long Short-Term Memory (LSTM) networks, have become the state-of-the-art \cite{sherstinsky2020fundamentals}. LSTMs are explicitly designed to model temporal dependencies, making them highly effective for financial and commodity time-series forecasting \cite{jha2019time}. Our work applies LSTMs in a novel context: not just to forecast prices, but to do so over a variable time horizon determined by the specific growth cycle of a recommended crop.

\subsection{Voice Interfaces for Inclusive Development (ICT4D)}
The field of Information and Communication Technologies for Development (ICT4D) has long recognized the potential of voice to overcome literacy barriers. Studies have consistently demonstrated that for low-literacy and novice users, voice-based interfaces are more usable, efficient, and trustworthy than their text-based counterparts \cite{sherwani2009speech}. Early systems were constrained by the rigid, menu-driven nature of Interactive Voice Response (IVR) technology. The recent proliferation of powerful, pre-trained models for Automatic Speech Recognition (ASR) and Text-to-Speech (TTS) has enabled a new generation of truly conversational interfaces \cite{radford2023robust}. Our system builds on this progress, adopting a strategic, asymmetric approach by fine-tuning an efficient ASR model for Kannada input and using a specialized, high-quality Kannada TTS model for output, adhering to HCI principles that prioritize clarity and trust for the end-user.

\section{System Architecture and Methodology}
\begin{figure*}[t]
  \centering
  \includegraphics[width=\textwidth]{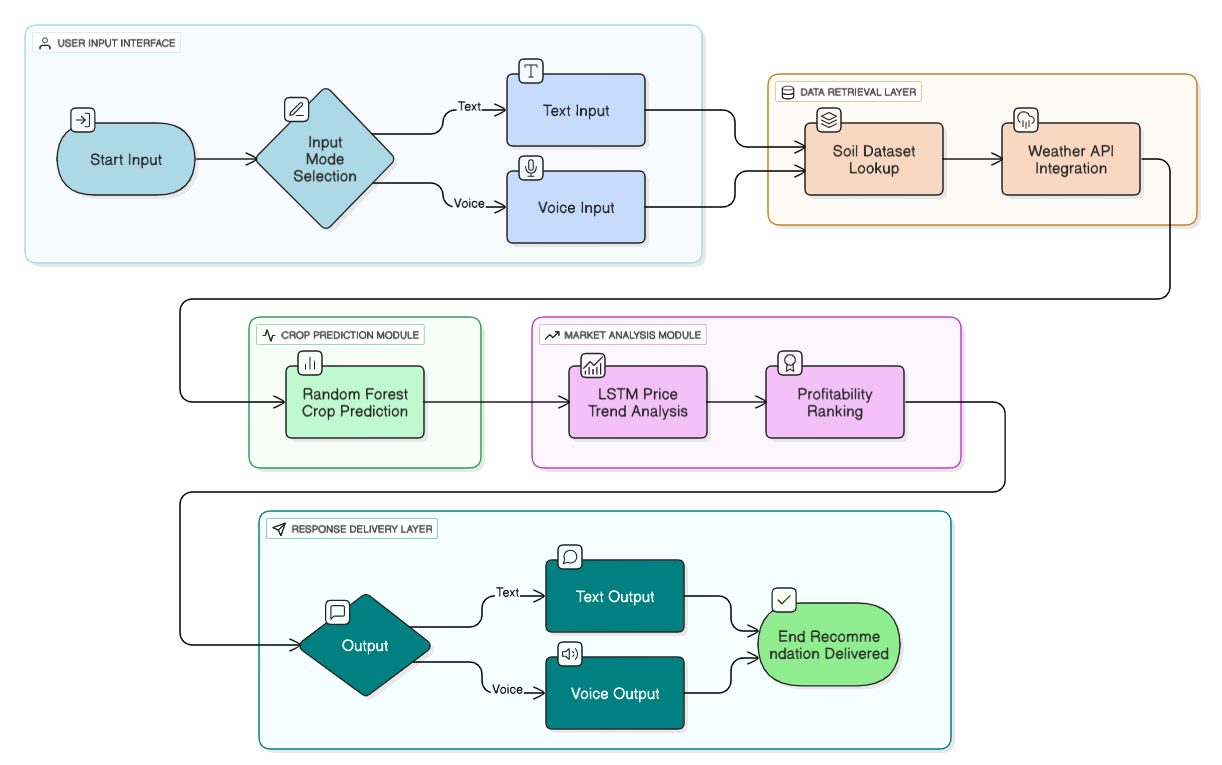}
  \caption{High-level architecture of the proposed system. A user's voice query in Kannada is transcribed by the ASR module. The intent is parsed, and the hybrid engine, which pulls real-time weather and soil data, executes its two-stage prediction. The final recommendation is synthesized into a spoken response by the TTS module.}
  \label{fig:architecture}
  \Description{A flowchart showing the system architecture. It starts with a user's voice input, which goes to an ASR module. The text output goes to a hybrid recommendation engine, which in turn queries data APIs. The engine's output goes to a TTS module, which produces a voice output for the user.}
\end{figure*}

The system is designed as a modular, end-to-end pipeline that processes a farmer's spoken query and returns a data-driven, voice-based recommendation. The architecture, shown in Figure \ref{fig:architecture}, consists of data ingestion and preprocessing, the core hybrid recommendation engine, and the voice interaction module.

\subsection{Data Sources and Preprocessing}
The system's predictive power is built upon four heterogeneous datasets, summarized in Table \ref{tab:datasets}. Preprocessing steps included mean imputation for any missing numerical values and feature scaling using scikit-learn's \texttt{MinMaxScaler} to normalize data ranges, ensuring that no single feature disproportionately influences model training. The geocoding of a user's address to a specific district is handled by first converting the address to latitude/longitude coordinates and then using a K-Nearest Neighbors (KNN) classifier (with k=1) trained on the district coordinate data to find the closest district centroid.

\begin{table}[h]
  \small 
  \caption{Summary of Datasets Used in the System}
  \label{tab:datasets}
  \centering
  \begin{tabular}{@{}lll@{}}
    \toprule
    \textbf{Dataset Name} & \textbf{Description} & \textbf{Key Features} \\
    \midrule
    \texttt{CROPP.csv} & Agronomic data & N, P, K, temp, pH, etc. \\
    \texttt{soil\_data1.csv} & District soil profiles & pH, N, P, K \\
    \texttt{district\_coords.csv} & Geospatial mapping & District, Lat, Lon \\
    \texttt{dataset2.csv} & Market prices & Crop, Date, Price \\
    \bottomrule
  \end{tabular}
\end{table}

\subsection{Hybrid Recommendation Engine}
The engine's intelligence lies in its two-stage predictive process.

\subsubsection{Stage 1: Agronomic Suitability Modeling}
This stage identifies a set of crops that are agronomically viable. The core is a Random Forest classifier \cite{breiman2001random}.  
Given a feature vector:
\[
X = 
\begin{bmatrix}
N, P, K, \text{temp}, \text{hum}, \text{pH}, \text{rain}
\end{bmatrix}
\]
the model predicts the probability \( P(C_i \mid X) \) for each crop \( C_i \). The model, implemented using scikit-learn's RandomForestClassifier, was configured with hyperparameters tuned via cross-validation (see Table~\ref{tab:hyperparams}). The top 3 crops with the highest probabilities are selected.

\subsubsection{Stage 2: Economic Profitability Forecasting}
This stage forecasts the future market price for each suitable crop. We employ an LSTM network. For each candidate crop, its historical price series is extracted and normalized. The data is transformed into sequences (look-back period = 6 months). The network architecture, implemented in TensorFlow/Keras, is detailed in Table \ref{tab:hyperparams}. The model is compiled using the 'adam' optimizer and 'mean\_squared\_error' loss function, with an \texttt{EarlyStopping} callback. To forecast for a harvest `n` months away, the model predicts iteratively.

\begin{table}[h]
  \caption{Model Hyperparameter Configuration}
  \label{tab:hyperparams}
  \begin{tabular}{@{}lll@{}}
    \toprule
    Model & Hyperparameter & Value \\
    \midrule
    Random Forest & n\_estimators & 100 \\
    & max\_depth & 20 \\
    & min\_samples\_leaf & 1 \\
    & criterion & 'gini' \\
    \midrule
    LSTM & LSTM Layer 1 Units & 64 \\
    & LSTM Layer 2 Units & 32 \\
    & Activation & ReLU \\
    & Dropout Rate & 0.2 \\
    & Optimizer & Adam \\
    & Loss Function & Mean Squared Error \\
    \bottomrule
  \end{tabular}
\end{table}

\subsection{Voice Interaction Module}
To ensure accessibility, the system uses a fully voice-based interface.
\begin{itemize}
    \item \textbf{Speech-to-Text (STT):} We use the distil-whisper/\-distil-large-v2 model, a distilled version of OpenAI's Whisper \cite{radford2023robust}. It has been fine-tuned on a corpus of Kannada speech data.
    
    \item \textbf{Text-to-Speech (TTS):} For output, we use the facebook/\-mms-tts-kan model, part of Meta’s Massively Multilingual Speech project. This model is selected for its high-fidelity Kannada output \cite{pratap2023scaling}.
\end{itemize}

\section{Results and Discussion}
The system was evaluated through component-wise model testing and an end-to-end case study analysis.

\begin{figure*}[t]
  \centering
  \includegraphics[width=\textwidth]{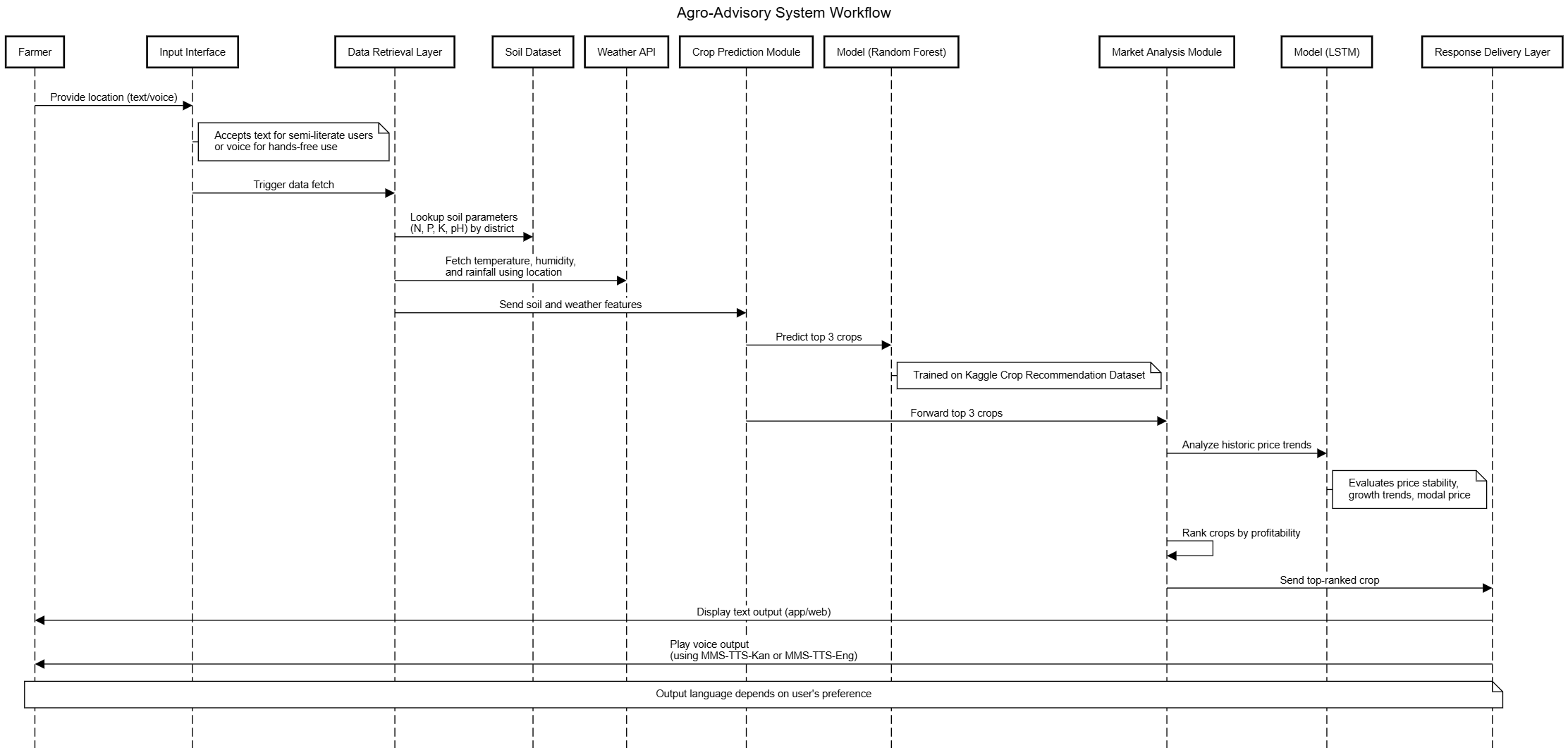}
  \caption{System-level interaction diagram illustrating the key components and their communication flow during the execution of a user query.}

  \label{fig:sequence_diagram}
  \Description{A sequence diagram illustrating how the user query flows from ASR, to intent parser, to the recommendation engine, and finally to TTS. External API calls fetch contextual data before generating the response.}
\end{figure*}

\subsection{Quantitative Model Performance}
The Random Forest classifier for crop suitability demonstrated excellent performance, achieving an overall accuracy of \textbf{98.5\%} on a 20\% held-out test set. Precision, recall, and F1-score for most crop classes were above 0.95.

The LSTM price forecasting model also showed strong performance. When evaluated on a one-month-ahead forecast task, the average Root Mean Squared Error (RMSE) was \rupee 3.50 per kg, and the Mean Absolute Percentage Error (MAPE) was approximately 5.8\%.

\subsection{Qualitative Case Study: End-to-End Recommendation}
To illustrate the system's practical utility, we present a case study for a user located in the Hassan district of Karnataka.
\begin{enumerate}
    \item \textbf{Input:} The user provides their location. The system geocodes it to Hassan and retrieves its soil data (N: 125, P: 29, K: 260, pH: 6.2) and real-time weather.
    \item \textbf{Stage 1 (Suitability):} The Random Forest model identifies 'Coffee', 'Pepper', and 'Maize' as the top 3 most suitable crops.
    \item \textbf{Stage 2 (Profitability):} The LSTM models forecast prices based on their respective growth periods:
    \begin{itemize}
        \item Coffee (9-month growth): Forecasted price of \textbf{\rupee 255/kg}.
        \item Pepper (6-month growth): Forecasted price of \rupee 480/kg.
        \item Maize (4-month growth): Forecasted price of \rupee 22/kg.
    \end{itemize}
    \item \textbf{Final Recommendation:} The system recommends \textbf{Pepper} as it has the highest predicted price. The spoken output in Kannada is generated.
\end{enumerate}
This case study highlights the system's core value: it moves beyond simple suitability to provide a financially optimized recommendation.
\begin{table*}[t]
  \footnotesize 
  \caption{Comparative Analysis of Speech Technology Choices and Rationale for Selection}
  \label{tab:speech_tech}
  \centering
  \renewcommand{\arraystretch}{1.1} 
  \begin{tabular}{@{}l l p{0.13\textwidth} p{0.23\textwidth} p{0.33\textwidth}@{}}
    \toprule
    Component & Model/Service & Type & Key Characteristics & Rationale in System \\
    \midrule

    \multirow{2}{*}{Input (ASR)} 
    & distil-whisper/distil-large-v2 
    & Self-hosted 
    & 6× faster, 49\% smaller than base Whisper. \newline Needs Kannada fine-tuning. 
    & Efficient, low-cost inference. \newline Supports scalable, serverless deployment. \\

    & Google Cloud Speech-to-Text 
    & API 
    & High accuracy, multilingual, managed service. 
    & Rejected: high and unpredictable long-term cost. \\

    \midrule

    \multirow{2}{*}{Output (TTS)} 
    & facebook/mms-tts-kan 
    & Self-hosted 
    & High-fidelity voice output optimized for Kannada. 
    & Preferred for clarity, naturalness, and user trust. \\

    & Google Cloud Text-to-Speech 
    & API 
    & Large selection of WaveNet voices. 
    & Rejected due to high per-character pricing. \\

    \bottomrule
  \end{tabular}
\end{table*}

\subsection{Limitations and Critical Discussion}
Despite promising results, the system has limitations. First, recommendation accuracy depends on the quality of input data. Inaccuracies in government-provided data will propagate through the pipeline. Second, ASR performance may vary across diverse Kannada dialects, requiring continuous fine-tuning. Third, the LSTM model assumes future trends will resemble historical patterns and may not capture black-swan events. A critical ethical consideration is the risk of promoting monocultures if the model consistently favors a single crop for a region. This could reduce biodiversity and increase vulnerability to pests and diseases.

\section{Conclusion}
This paper has presented a hybrid machine learning framework designed to provide accessible, economically optimized agricultural advice. By integrating a Random Forest model for agronomic suitability with an LSTM network for market price forecasting, our system delivers recommendations that are both viable and profitable. The novel use of a voice-first interface makes this powerful analytical tool accessible to low-literacy users.

The results confirm the efficacy of our approach. This work serves as a robust proof-of-concept for a new class of decision support systems that are not only technologically advanced but also deeply human-centered.

Future work will proceed along three main trajectories. First, we will enhance the engine by incorporating a cost-of-cultivation model to provide net profit forecasts and a diversification model to mitigate the risk of promoting monocultures. Second, we plan to implement a federated learning architecture to allow for model personalization without compromising user privacy \cite{konevcny2016federated}. Finally, we will conduct a large-scale randomized controlled trial (RCT) to rigorously measure the causal impact of our system on farmer income and economic well-being.

\bibliographystyle{ACM-Reference-Format}

\balance 

\end{document}